\documentclass[nohyperref]{article}

\usepackage{microtype}
\usepackage{graphicx}
\usepackage{subfigure}
\usepackage{booktabs} %
\usepackage[T1]{fontenc}
\usepackage{float}
\usepackage[utf8]{inputenc}

\usepackage{hyperref}
\usepackage{listings}

\usepackage{multirow}
\usepackage[table]{xcolor}

\usepackage{iclr2023_conference,times}
\iclrfinalcopy
\usepackage{amsmath}
\usepackage{amssymb}
\usepackage{mathtools}
\usepackage{amsthm}

\usepackage[capitalize,noabbrev]{cleveref}

\theoremstyle{plain}

\theoremstyle{definition}

\theoremstyle{remark}

\usepackage[textsize=tiny]{todonotes}

\title{Learning Translation Quality Evaluation on Low Resource Languages from Large Language Models}

\usepackage[compact]{titlesec} 
\titlespacing*{\section}{14pt}{7pt}{4pt}
\titlespacing*{\subsection}{6pt}{3pt}{1pt}
\author{Amirkeivan Mohtashami \thanks{Work done during an internship at Google.}\\
Department of Computer Science, EPFL \\
Lausanne, Switzerland \\
\texttt{amirkeivan.mohtashami@epfl.ch} \\
\And
Mauro Verzetti \& Paul K.~Rubenstein \\
Google \\
Zurich, Switzerland \\
\texttt{\{verzetti, paulrubenstein\}@google.com} \\
}

\newif\ifdatareleased
\datareleasedfalse
\newcommand{\datarel}[1]{\ifdatareleased#1\fi}
\begin{document}
\maketitle

\begin{abstract}
Learned metrics such as BLEURT have in recent years become widely employed to evaluate the quality of machine translation systems.
Training such metrics requires data which can be expensive and difficult to acquire, particularly for lower-resource languages. 
We show how knowledge can be distilled from Large Language Models (LLMs) to improve upon such learned metrics without requiring human annotators, by creating synthetic datasets which can be mixed into existing datasets, requiring only a corpus of text in the target language.
We show that the performance of a BLEURT-like model on lower resource languages can be improved in this way.
\end{abstract}
\section{Introduction}

A machine translation system is typically evaluated by comparing its output on a given input sentence with one made by a professional translator. 
Concretely, a test set is constructed by taking a corpus of sentences $\{a_1, a_2, \ldots, a_n\}$ in language A and having a human translate them to language B $\{b_1, b_2, \ldots, b_n\}$. 
A machine translation system $T: A \rightarrow B$ is then evaluated by comparing $T(a_i)$ and $b_i$ using some metric $m\left(T(a_i), b_i\right)$.

Until recently, commonly used metrics such as BLEU \citep{papineni2002bleu} and ROGUE \citep{lin2004rouge} were generally based on number of co-occurring n-grams.
Advantages of such methods include that they are easy to interpret, do not require learning from data, and have been shown to generally correlate with human judgement when averaged over a corpus of sentences.

Nonetheless, these approaches fail when sentences are semantically similar but differ significantly in phrasing.
For example, although the sentences \emph{the sky is clear} and \emph{there are no clouds above} have no words in common, their meaning is similar and a good metric should assign a high score to such pairs as well.
To alleviate this issue, recent works such as BLEURT \citep{Sellam2020-gy} and COMET \citep{Rei2020-ac} have considered neural-network based metrics which work based on semantic similarity, but come at the cost of decreased interpretability and being more computationally intensive to run.

Such metrics can be learned by treating the problem of defining a metric $m\left(T(a_i), b_i\right)$ as a regression problem, where a high score should be returned if $T(a_i)$ and $b_i$ are similar, and a low score if they are different.
This requires an annotated dataset of reference and candidate sentences, along with scores provided by human annotators.
Phrasing the problem this way, $m$ is not a translation metric \textit{per se} but rather a \textit{sentence similarity} metric which can be used to evaluate translation quality.

Although some public datasets for this purpose exist --- notably the WMT datasets released each year as part of the WMT Machine Translation Evaluation task \citep{Ma2019-nq} --- it is in general expensive to create such datasets as they require expensive annotation by skilled humans. 
As such, available datasets tend to be either very small or restricted to popular languages, and therefore the performance of metrics trained with these datasets may deteriorate on lower resource languages.

Fortunately, recent developments in large language models (LLMs) allow for new possible solutions for many tasks including this one. Due to being trained on internet-scale textual datasets, LLMs such as PaLM \citep{Chowdhery2022-ju} and GPT-3 \citep{Brown2020-lm} have knowledge of a large number of languages; the training sets of both PaLM and GPT-3 include more than 100 languages. Furthermore, these models have demonstrated a remarkable ability to perform a wide variety of tasks by selecting prompts with a few (or even zero) examples.

As such, LLMs could be used directly as a metric to evaluate machine translation systems by asking the model to numerically rate how similar sentence pairs are: Table~\ref{tab:palm-vs-bleurt} shows the correlation between the human scores and those generated by PaLM using the method described in Section~\ref{subsec:scoring-pairs} which can be seen is in par with the correlation of the scores generated by the BLEURT-like baseline.
While this would be significantly cheaper and faster than doing the same with professional humans, cost and time may still be prohibitive since inference with such models requires significant compute and is slow compared to standard metrics. 

Therefore, we propose another approach to use LLMs to create datasets of sentence pairs with similarity ratings, which can be used to train smaller neural metrics such as BLEURT. 
This means that the cost of querying the LLM is incurred only once when the dataset is generated, and inference times and cost of the resulting metric are unchanged.
The LLM is used to create datasets for languages for which such data is scarce or non-existent.
In doing so, knowledge of lower resource languages is distilled from the LLM into the neural metric.

The contributions of this work are:
\begin{itemize}
    \item We demonstrate that LLMs can generate synthetic datasets for training sentence similarity metrics.
    \item We use this to construct to low resource language training data for translation quality metrics, and demonstrate an improvement over prior art by training a BLEURT-like model with this additional data.
\end{itemize}

\begin{table*}[t]
\centering
\begin{tabular}{|c|c|c|c|c|}
    \hline
          & WMT & \multicolumn{3}{|c|}{Internal} \\\hline
          & English & Arabic & Urdu & Mongolian \\\hline
          SentenceBLEU & 0.316 & 0.061 & 0.069 & 0.025 \\\hline
         BLEURT-like baseline & 0.592(0.001) & 0.210(0.001) & 0.255(0.001) & 0.123(0.001) \\\hline
         PaLM with scoring prompt &0.541 & 0.163 & 0.262 & 0.105 \\\hline
\end{tabular}
\caption{Spearman's $\rho$ correlation between scores given by human raters and automatic raters. This demonstrates that given the right prompting, PaLM can be used to directly score sentence similarity. For the baseline the average over 5 fine-tuning is reported along with the standard error in parenthesis. See Section~\ref{sec:models-and-data} for details about the baseline. SentenceBLEU refers to the sacrebleu reference implementation \citep{post-2018-call}}
\label{tab:palm-vs-bleurt}
\end{table*}

\section{Related Works}
\label{app:related-works}
In recent years, language models have considerably grown in size leading to a remarkable increase in their capabilities and performance. GPT-3 \citep{Brown2020-lm} and PaLM \citep{Chowdhery2022-ju} are current examples of the state of the art with 175B and 540B parameters respectively.
These models are trained in an unsupervised manner over large corpus of text in a variety of languages, contexts, and settings.
In this paper we work with PaLM, though any LLM could be used.
(In particular, further advances in LLMs made made in the future may translate into further gains in performance.)

Various learned sentence similarity metrics have been proposed in the literature.
BERTscore \citep{Zhang2019-ab} uses a pre-trained BERT model \citep{Devlin2018-ov} to obtain token embeddings and uses the cosine similarity between the embeddings of two sentences to compute their similarity. 
Several recent methods also build upon pre-trained language models but directly fine-tune them to solve the sentence similarity task as a regression.
For example, BLEURT \citep{Sellam2020-gy} fine-tunes BERT first over synthetic data labeled using traditional metrics such as BLEU \citep{Papineni2002-zh}, followed by another fine-tuning over human labeled data. In a follow-up \citet{sellam2020learning} note that the initial fine-tuning for multi-lingual scenarios is not as benefical and can be ommitted. COMET \citep{Rei2020-ac} is another learned metric for translation quality evaluation that also takes into account the sentence in the source language. To avoid the need for human-rated sentence pairs, \citet{thompson2020automatic} train PRISM in an unsupervised manner to estimate how much a sentence can be considered a para-phrased version of another sentence. Similarly, BARTScore \citep{yuan2021bartscore} uses a pre-trained language model to estimate the probability of generating one sentence to continue another sentence and uses the estimated value as a surrogate for their similarity.

Several recent works also propose methods that are applicable to any learned metric and can be used to improve their performance.
For example \citet{Pu2021-yi} propose to train a smaller student model using a large amount of unlabeled data over the specific set of target languages in order to both improve the speed and the accuracy over those languages. %

Existing works have explored the use of  LLMs to generate synthetic datasets.
\citet{Schick2021-hk} also consider generating data to train sentence similarity metrics, but only for English and not for application to translation quality evaluation.
\citet{Rosenbaum2022-km} use LLMs to generate structured data and a corresponding expression in natural language; while they do consider a multilingual setting, they are restricted only to popular languages.

The work presented here is unique in two regards: 
first, we demonstrate the capability of LLMs to generate datasets for lower-resource languages; 
and second we apply this to learning sentence similarity metrics for application to translation quality evaluation.
Given the near-universal lack of data for lower-resource languages, our approach to transfer this knowledge from LLMs can be useful to many tasks, even those already having a large amount of data in high resource languages.

\section{Dataset Generation}
\label{sec:dataset-generation}

Our method to generate datasets assumes access only to a corpus of text in the target language and access to a suitable LLM.
In our experiments we use the multilingual C4 dataset \citep{Raffel2019-ep} for this purpose, however any other available sources could instead be used (books, news, government documents or transcriptions from parliaments as examples).

We generate datasets in four steps. 
In the first step, we select sentences from the text corpus. Since the corpus contains paragraphs extracted from the web, there are many segments which do not constitute a complete sentence such submission date, number of comments, etc. We apply basic filtering to remove such segments.
In the second step, we query PaLM to generate a second sentence derived from this, creating a sentence pair.
In the third step, we query PaLM to score each of the sentence pairs generated in the first step according to the similarity of the sentences.
Finally, we apply heuristic filtering to the generated triples to remove lower quality data.

\subsection{Selecting Sentences from Corpus}
For each record in multilingual C4 dataset \citep{Raffel2019-ep} we extract the first sentence from each line in that record. We use an internal sentence segmentation model to detect sentence boundaries. Subsequently, we filter out sentences that do not start with an alphabet character or do not finish with a punctuation mark. 
The resulting set of sentences are fed through the pipeline to generate sentence pairs and then score the generated pairs. \datarel{For the released validation sets, the records were instead picked from Wikipedia dataset.}

\subsection{Generating Sentence Pairs}
To generate the sentence pair, we use PaLM to introduce alterations in a sentence picked from our text corpus.
Since the final model is a metric for evaluating high-quality translation systems, we focus on generating altered sentences that could be outputs of such systems. Therefore, we focus the main instruction given to PaLM on the meaning of the generated alteration. In particular, we ask PaLM to make alterations to achieve different objectives such as rephrasing the sentence or changing its meaning as determined by the prompt. While the instruction does not prohibit adding grammatical mistakes to the sentence, however PaLM usually generates a grammatically correct sentence.\looseness=-1 

We also want the generated dataset to be diverse. Thus, to avoid biasing PaLM to a specific change, e.g. only replacing a word with its synonym, we use a zero-shot prompt, i.e. we describe the task without providing examples.

Similar to prior works \citep{Brown2020-lm,Chowdhery2022-ju}, we create a template with a placeholder for the input sentence and use it to query PaLM.
The output is generated with deterministic \textit{zero-temperature decoding}, whereby the most probably next token is iteratively chosen.
While testing various templates, we observed that PaLM sometimes rephrases the given task in terms similar to competitive programming tasks.
We utilize this observation and directly phrase our task in a competitive programming task template.
Since these tasks usually contain various constraints explained in the description, using this template allows explaining the objective clearly and in our experience reduces the chances of the model ignoring the given constraints.

We found it effective to follow chain-of-thought reasoning \citep{Chowdhery2022-ju} by requesting an explanation of the changes to be made before producing the final sentence.
In the absence of this, we observed that in many cases the generated sentences did not satisfy the given constraints.
When not asking for an explanation, we frequently observed that the model output follows the generated sentence with a description that falsely asserts that the generated sentence satisfies the given constraints.
This description hints that the model has not ignored the constraints completely but fails to apply them when generating the sentence when not asked for a concrete explanation in advance.

We tested various prompts and based on the number of successfully generated pairs, settled on the following prompts to generate our final dataset:

\paragraph{Prompts for Preserving the Meaning} We ask PaLM to \textit{re-write a sentence without using any of the words in the original sentence while preserving the meaning}. For the explanation, we either ask it to provide a list of five differences between the input and the output or ask it to provide three ideas used to change the input into the output.

\paragraph{Prompts for Changing Meaning} In this template, we ask PaLM to \textit{change a small number of words in the original sentence to significantly change the meaning but keeping the context}. Similar to the first template, we ask for a list of five differences to force the model to provide an explanation of the output.

Appendix Figure~\ref{fig:generation-template} shows the generic structure of the templates described above.

\subsection{Scoring Sentence Pairs}\label{subsec:scoring-pairs}
After generating a sentence pair, we query PaLM a second time and use it to score the similarity of the two sentences in the pair.
Note that a score can also be guessed based on the template that was used for generation. For example, the template can instruct PaLM to apply only minor changes. However, we noticed that PaLM can violate such instructions quite often, making the guessed score to be less reliable. Therefore, we decided to obtain a more accurate score by querying PaLM again.

Similar to generation, we use a template with two placeholders (one for each sentence in the pair) to query PaLM.%
However, we follow a different approach to obtain the score than greedy decoding.

The template defines a discrete set of allowed scores: the integers $\{0,1,\ldots, 4\}$.
We end the template such that the score would be the natural continuation and compute the probability of each score as a continuation.
Next, we take a weighted average of these scores, with weighting given by these probabilities. 
This approach allows obtaining the score in a continuous scale instead of a discrete one and is more reliable as it can take into account uncertainties of the model between two or more scores.\looseness=-1 

For this task, we tested several templates including both zero-shot and few-shot templates, measuring the performance of each template according to the correlation metrics between the generated scores and human scores over a subset WMT Shared Metric Task dataset \citep{Ma2019-nq}.
The final template was a few-shot template including an explanation of the task and the grading scale, followed by examples from several languages.
The general structure of this template can be seen in Appendix Figure~\ref{fig:scoring-template}.
The correlation of the generated scores using this template with human scores over can be seen in Table~\ref{tab:palm-vs-bleurt}.

\subsection{Final Cleaning}
To ensure the quality of the generated sentence pairs, we select a subset of the generated sentence pairs based on some heuristics:

\begin{itemize}
    \item \textbf{Sentence Length}: We ensure the reference sentence is neither too long or too short. In particular, we filter out any sentence below 20 characters and above 300 characters.
    \item \textbf{Generated/Reference Length Ratio}: We ensure that the generated sentence is not much shorter or much longer than the reference. In particular, we allow pairs where the ratio of the generated sentence's length to the original sentence's length is at least $0.8$ and at most $2$.
    \item \textbf{Minimum Edit Distance}: We ensure the two sentences have at least edit distance $5$ in order to avoid too similar pairs.  Note that this condition is relaxed enough that most pairs with a noticeable change pass it. Thus, this condition mainly filters cases where there are almost no changes, e.g. when the only difference is a removed space. While such pairs could also be useful for training, the ratio of such pairs among the pairs generated by PaLM is quite high which can harm the quality of the final dataset. To avoid this, we filter all such pairs from PaLM output. Note that pairs with identical sentences are already present in the baseline training data as explained in more detail in Section~\ref{sec:models-and-data}.
\end{itemize}

\section{Models, Data and Experiments}
\label{sec:models-and-data}
\paragraph{Model} We train a BLEURT-like model, taking as a starting point the mT5-XL  model with 3.7B parameters\footnote{Available at \url{https://github.com/google-research/multilingual-t5}} which is pretrained using the multilingual C4 dataset \citep{Raffel2019-ep}.
Similar to \citet{sellam2020learning} we omit the "mid-training" step proposed in the original BLEURT paper and fine-tune pre-trained model directly on rated sentence pairs.

\paragraph{Baseline Training Data} We obtained the dataset used to train the BLEURT-20\footnote{For additional details see details available at  \url{https://github.com/google-research/bleurt}} model from the authors which consists of the WMT Metrics Shared Task data with additional augmentations.

\paragraph{Additional LLM-synthesized Training Data}

We generate additional datasets in several languages by following the process described in Section \ref{sec:dataset-generation}.
Depending on the experiment, a combination of these datasets are used as the final training set.

\paragraph{Test Data}
The lack of data for lower-resource languages extends to the test data as well, making it hard to evaluate new methods.
In this work, we report our result over a small internal dataset which contains a set of sentences translated by human experts and different machine translation systems from English to different target languages, with human-assigned scores of translation quality obtained by comparing the human reference translations with the machine translations.
The test set for each language contains 1-2k examples.
We note that in this dataset each sentence pair is rated only by a single rater, which may be a source of label noise.
We expect a good sentence similarity model to correlate with the human ratings and therefore, use the correlation between our model's score and the human scores as a metric.

\newcommand{\includedcell}{\cellcolor{green!25}}
\begin{table*}[t]
    \centering
    \begin{tabular}{|c|c|c|c|c|c|c|}
         \hline
         Dataset&Language&BLEU&Baseline&Small&Medium&Large\\\hline
         WMT&All&0.287&\includedcell0.573(0.001)&\includedcell0.573(0.000)&\includedcell0.573(0.002)&\includedcell\textbf{0.576(0.001)}\\\hline
\multirow{7}{*}{Internal}&Spanish&0.130&\textbf{0.321(0.001)}&\includedcell0.316(0.001)&\includedcell0.312(0.001)&\includedcell0.315(0.002)\\\cline{2-7}
&Mongolian&0.025&0.142(0.002)&\includedcell0.152(0.005)&\includedcell0.149(0.004)&\includedcell\textbf{0.156(0.003)}\\\cline{2-7}
&Amharic&0.035&0.273(0.003)&\includedcell0.289(0.002)&\includedcell\textbf{0.300(0.004)}&\includedcell\textbf{0.300(0.006)}\\\cline{2-7}
&Urdu&0.069&0.261(0.001)&0.260(0.002)&\includedcell0.264(0.002)&\includedcell\textbf{0.279(0.003)}\\\cline{2-7}
&Belarusian&0.051&\textbf{0.214(0.003)}&0.212(0.005)&\includedcell0.201(0.006)&\includedcell0.199(0.005)\\\cline{2-7}
&Punjabi&0.080&0.184(0.004)&0.171(0.002)&\includedcell0.185(0.004)&\includedcell\textbf{0.194(0.006)}\\\cline{2-7}
&Macedonian&0.159&0.265(0.004)&\textbf{0.278(0.004)}&\includedcell0.268(0.004)&\includedcell0.266(0.003)\\\cline{2-7}
&Persian&0.150&0.418(0.001)&0.424(0.001)&0.421(0.002)&\includedcell\textbf{0.425(0.002)}\\\cline{2-7}
&Arabic&0.061&0.229(0.001)&0.229(0.001)&0.225(0.001)&\includedcell\textbf{0.236(0.001)}\\\hline
    \end{tabular}
    \caption{Pearson's R correlation between the true score and the output of models obtained from different sets of experiments. For each set, 5 models are trained and the average is shown with standard error in parenthesis. BLEU refers to the sacrebleu sentenceBLEU reference implementation \citep{post-2018-call}. A colored cell at the intersection of a language and an experiment indicate that the experiment includes the language.}
    \label{tab:exp-results}
\end{table*}

\paragraph{Experiments}
We generate a small dataset for multiple languages. 
The languages used along with the size of the generated dataset is listed in Table~\ref{tab:dataset-details}.
We perform 3 set of experiments, adding a subset of the generated datasets to the baseline dataset. In particular, the training data for the experiment sets are as follows (refer to Section~\ref{sec:models-and-data} for more details on the baseline dataset):
\begin{itemize}
\item Small: Bsaeline data plus generated data for Spanish, Mongolian, and Amharic.
\item Medium: Small experiment data plus generated data for Urdu, Belarusian, Punjabi, and Macedonian.
\item Large: Medium experiment data plus generated data for Arabic, and Persian.
\end{itemize}
In Appendix~\ref{app:dataset-details}, Table~\ref{tab:dataset-details} shows which languages were included in each set. 

We measure the correlation between the trained model's score and the true score.
For the generated datasets, the true score is also generated by PaLM. The results are listed in Table~\ref{tab:exp-results}.
Comparing the correlation between the large experiments and the baseline, it can be seen that adding the generated data helps improve the final performance across many languages.
Furthermore, it can be seen that adding data from more languages is important.
For example, performance on Urdu only improves after adding the Urdu dataset.
Still, the performance on some languages such as Amharic keeps improving as more data from other languages is added. 

We note that it is not always the case that adding our data in some language improves the results for that language.
This may be because of a distribution shift between the test data and our synthesized data. 
It may also be due to label noise in our generated data.
Nonetheless, we see a broad trend of improvement as our data is added.

\section{Future Work \& Conclusion}
Our results demonstrate the efficacy of using large language models to generate datasets for lower-resource languages.  While we focus our attention on the task of sentence similarity in this work, the same method can be applied universally to other tasks facilitating the possibility of supporting more languages. Some examples are summarization and sentiment detection. 

Our method can also be further improved, for example by finding better prompts. One possible approach is to use automatic prompt engineering which could allow a more tuned prompt than the manually crafted prompts we used. 

Finally, with the advancement of large language models, it is now possible to use them to perform tasks without the need for a formal definition. We demonstrate one such example in this work by asking PaLM to generate a modified version of a given sentence. This ability could make it possible to automate other tasks that are hard to describe formally. 

\section*{Acknowledgement}
This work was done while Amirkeivan Mohtashami was doing an internship at Google. We acknowledge and thank Google for supporting this work as well as providing the necessary infrastructure to perform the experiments. We would like to thank Aditya Siddhant, Thibault Sellam, Lotem Golany, Michael Tschannen, Duc Dung Nguyen and Alex Tudor for support, feedback and helpful discussions over the course of this work. 
\clearpage
\bibliography{references}
\bibliographystyle{iclr2023_conference}
\clearpage
\appendix
\onecolumn

\section{Other Correlation Metrics}
There are various metrics to measure correlation of two sequences. We have already reported the correlation of the trained models with the true score using Pearson's R metric. Here, we additionally report the correlation as measured by Spearman's $\rho$ and Kendall's $\tau$ for compeleteness. %

\begin{table}[H]
    \centering
    \begin{tabular}{|c|c|c|c|c|c|}
         \hline
         Dataset&Language&Baseline&Small&Medium&Large\\\hline
         WMT&All&0.592(0.001)&0.591(0.001)&0.590(0.002)&\textbf{0.593}\\\hline
\multirow{7}{*}{Internal}&Spanish&\textbf{0.266(0.001)}&0.265(0.001)&0.260(0.001)&0.265\\\cline{2-6}
&Mongolian&0.123(0.002)&0.137(0.003)&0.134(0.002)&\textbf{0.140}\\\cline{2-6}
&Amharic&0.287(0.003)&0.295(0.002)&\textbf{0.303(0.004)}&0.295\\\cline{2-6}
&Urdu&0.255(0.001)&0.249(0.002)&0.245(0.003)&\textbf{0.260}\\\cline{2-6}
&Belarusian&\textbf{0.174(0.004)}&0.171(0.006)&0.151(0.005)&0.152\\\cline{2-6}
&Punjabi&0.116(0.003)&0.121(0.002)&0.126(0.005)&\textbf{0.150}\\\cline{2-6}
&Macedonian&0.300(0.002)&\textbf{0.304(0.002)}&0.297(0.002)&0.302\\\cline{2-6}
&Persian&0.415(0.001)&\textbf{0.417(0.002)}&0.414(0.002)&0.416\\\cline{2-6}
&Arabic&0.210(0.001)&0.210(0.002)&0.208(0.001)&\textbf{0.213}\\\hline

    \end{tabular}
    \caption{Spearman's $\rho$ correlation between the true score and the output of models obtained from different sets of experiments. For each set, 5 models are trained and the average is shown with standard error in parenthesis.}
    \label{tab:exp-results-spearman-rho}
\end{table}

\section{Structure of Prompt Templates}
\begin{figure}[H]
    \centering
    \begin{minipage}{\linewidth}
    \small
    \begin{lstlisting}
Abracadabra is the art of re-writing a sentence without using any of the words in the original sentence while preserving the meaning. For example a beautiful Abracadabra for "Hello" can be "Hi". A weak Abracadabra for "Hello" is "Bye". Write a program to generate an Abracadabra for a given sentence.

Input
The input contains contains a single line which contains a sentence in <LANGUAGE>.

Output
Output a single line containing the <LANGUAGE> Abracadabra for the given sentence in <LANGUAGE>. 

Sample Sections:
A) Sample Input
B) Five Differences Between the Sample Input Sentence and the Output Sentence
C) Sample Output
D) Additional Explanation

A) Sample Input
<SENTENCE>

B) Five Differences Between the Sample Input Sentence and the Output Sentence
\end{lstlisting}
    \end{minipage}
    \caption{Generic structure of the template for the prompt used to generate a modified sentence with the same meaning. Similar template is used with a different definition to obtain a modified sentence with a different meaning. The term \texttt{<SENTENCE>} and \texttt{<LANGUAGE>} are replaced by the sentence to be modified and its language respectively.}
    \label{fig:generation-template}
\end{figure}

\begin{figure}[H]
    \centering
    \begin{minipage}{\linewidth}
    \small
    \begin{lstlisting}
Task: Consider the pair of sentences and rate them on a scale of 0 to 4 based on how similar they are, where 0 is least similar and 4 is most similar. The sentences could be in any language.

An explanation of the ratings:
- 0: the hypothesis is incoherent, contains serious grammatical errors, words in different languages, or has meaning completely different from the reference.
- 1: the hypothesis has some relation to the reference, but has many errors.
- 2: the hypothesis has similar meaning to the reference, but contains grammatical errors.
- 3: the hypothesis has almost the same meaning as the reference, and is mostly grammatically correct.
- 4: the hypothesis has the same or very similar meaning as the reference, and is grammatically correct.

Reference: Xin chào tên bạn là gì?
Hypothesis: Xin chào tên bạn là gì?
Rating: 4

Reference: The dataset was too large to load into memory.
Hypothesis: All of the information together was impossible to load into memory.
Rating: 2

<ADDITIONAL_EXAMPLES>

Reference: <SENTENCE_1>
Hypothesis: <SENTENCE_2>
Rating: 
\end{lstlisting}
    \end{minipage}
    \caption{Generic structure of the template for the prompt used to score a sentence pair. In the zero-shot setting only the explanation and no examples are provided, though in practice we found providing more examples in a variety of languages gave the best results.}
    \label{fig:scoring-template}
\end{figure}

\section{Generated Training Data Details}
\label{app:dataset-details}
\begin{table}[H]
    \centering
    \begin{tabular}{|c|c|c|c|c|c|}\hline
        Dataset&Size ($\times$1000)&Baseline&Small&Medium&Large\\\hline
        Baseline&595&\checkmark&\checkmark&\checkmark&\checkmark\\\hline
        Spanish&15&&\checkmark&\checkmark&\checkmark\\\hline
        Mongolian&4&&\checkmark&\checkmark&\checkmark\\\hline
        Amharic&4&&\checkmark&\checkmark&\checkmark\\\hline
        Urdu&9&&&\checkmark&\checkmark\\\hline
        Belarusian&7&&&\checkmark&\checkmark\\\hline
        Punjabi&2&&&\checkmark&\checkmark\\\hline
        Macedonian&9&&&\checkmark&\checkmark\\\hline
        Arabic&4&&&&\checkmark\\\hline
        Persian&20&&&&\checkmark\\\hline
    \end{tabular}
    \caption{List of various datasets used for each set of experiments. The dataset sizes are rounded down to the nearest thousand.}
    \label{tab:dataset-details}
\end{table}
\end{document}